\documentclass[10pt,twocolumn,letterpaper]{article}

\usepackage{cvpr}
\usepackage{microtype}
\usepackage{times}
\usepackage{epsfig}
\usepackage{graphicx}
\usepackage{amsmath}
\usepackage{amssymb}
\usepackage{array}
\usepackage{subcaption}

\usepackage[labelformat=simple]{subcaption}

\usepackage[pagebackref=true,breaklinks=true,letterpaper=true,colorlinks,bookmarks=false]{hyperref}

  \cvprfinalcopy 

\def\cvprPaperID{2601} 

\ifcvprfinal\fi
\begin{document}

\title{People, Penguins and Petri Dishes: Adapting Object Counting Models To New Visual Domains And Object Types Without Forgetting}

\author{Mark Marsden$^1$, Kevin McGuinness$^1$, Suzanne Little$^1$, Ciara E. Keogh$^2$, Noel E. O'Connor$^1$\\
$^1$Insight Centre for Data Analytics \\
Dublin City University, Ireland\\
{\tt\small mark.marsden@insight-centre.org \{kevin.mcguinness,suzanne.little,noel.oconnor\}@dcu.ie}
\\
$^2$Conway Institute, School of Medicine\\
University College Dublin, Ireland\\
{\tt\small ciara.keogh@ucdconnect.ie}
}

\maketitle

\begin{abstract}
 In this paper we propose a technique to adapt a convolutional neural network (CNN) based object counter  to additional visual domains and object types while still preserving the original counting function. Domain-specific normalisation and scaling operators are trained to allow the model to adjust to the statistical distributions of the various visual domains.  The developed adaptation technique is used to produce a singular patch-based counting regressor capable of counting various object types including people, vehicles, cell nuclei and wildlife.  As part of this study a challenging new cell counting dataset in the context of tissue culture and patient diagnosis is constructed. This new collection, referred to as the Dublin Cell Counting (DCC) dataset, is the first of its kind to be made available to the wider computer vision community. State-of-the-art object counting performance is achieved in both the Shanghaitech (parts A and B) and Penguins datasets while competitive performance is observed on the TRANCOS and Modified Bone Marrow (MBM) datasets, all using a shared counting model. 
\end{abstract}

\section{Introduction}
\begin{figure}

\begin{minipage}{\textwidth}

\begin{subfigure}{\textwidth}
\includegraphics[width=0.23\textwidth]{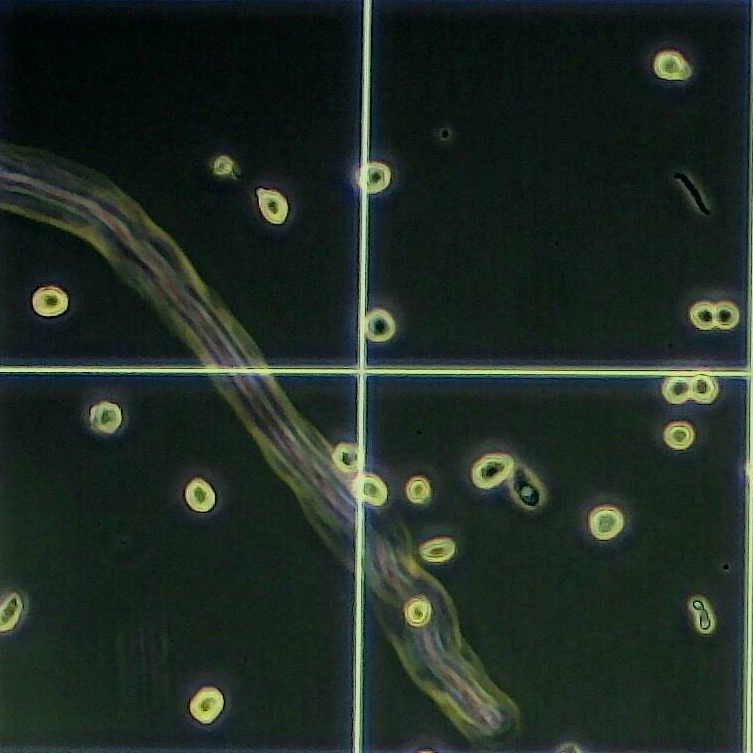}
\space
\includegraphics[width=0.23\textwidth]{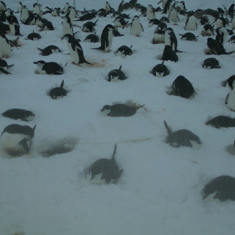}
 
\end{subfigure}
\hfill
\begin{subfigure}{\textwidth}

\vspace{0.1cm}

\includegraphics[width=0.23\textwidth]{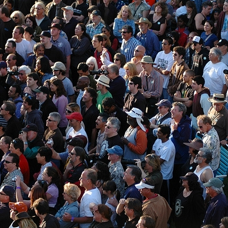}
\space
\includegraphics[width=0.23\textwidth]{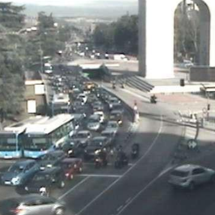}
\end{subfigure}
\end{minipage}
\hfill
\caption{Examples of the challenging images encountered in different object counting domains.}
\label{examples}
\end{figure}

Vision-based object counting is an important analysis step in many observational scenarios including wildlife studies, microscopic imaging and CCTV surveillance. An accurate object counting system can provide valuable insights such as the congestion level of a public square (crowd counting), the  level of traffic on a motorway (vehicle counting), the inferred migration patterns of a penguin colony (wildlife counting) or the proliferation of  cancerous cells in a patient (cell counting). There is a common set of challenges for vision-based object counting which limit counting accuracy in all of these domains. These challenges include object scale and perspective issues, visual occlusion and poor illumination. To date, these highly related counting tasks have been tackled separately with algorithms engineered specifically to perform object counting in a given domain. Examples of these challenges are shown across several visual domains in Figure \ref{examples}.

Vision based object counting has been tackled using both object detection \cite{lin2001estimation,wu2005detection,ge2009marked}  and count regression techniques \cite{change2013semi,chen2012feature,lempitsky2010learning,liu2014robustness,chan2012counting} with both types of approach having their own specific weakness (visual occlusion and model overfitting respectively). The utilisation of convolutional neural networks (CNN) and hardware accelerated optimisation has lead to significant improvements in the accuracy of regression-based counting approaches across several visual domains \cite{Zhang2015,onoro2016towards,cohen2017count}.  This paper looks to build upon this work and investigate if a patch-based object count regressor can be adapted to additional object types and visual domains while maintaining accuracy in the original counting task. The potential benefits of a modular counting architecture include transfer learning and the removal of redundant model parameters. The significant variation in statistical distribution between visual domains is one of the main obstacles to domain adaptation in computer vision \cite{rebuffi2017learning}. This challenge is addressed in our approach by including a set of domain-specific scaling and normalization layers  distributed throughout the network which make up only a small fraction of the overall parameter count \cite{rebuffi2017learning}. A sequential training procedure inspired by the work of Rebuffi \etal \cite{rebuffi2017learning} allows the network to be extended to new counting tasks over time while preserving all previously learned counting functions. The set of count estimates produced using the proposed patch-based counting model are then refined using  an efficient fully convolutional neural network which utilises the wider scene context to mitigate possible count errors. The proposed framework is also extended to perform a visual domain classification in the event of the observed domain being unknown, highlighting the versatility and modular nature of the approach. The core contributions of this paper can be summarised as follows:

\begin{itemize}
\item An extendable CNN architecture for patch-based, multi-domain object counting is developed for the first time;
\item A fully convolutional neural network is utilised to refine a set of count estimates for an entire image;
\item A challenging, representative dataset for cell counting in a tissue culture/patient diagnosis setting is proposed;
\item Domain adaptation in object counting is shown to produce more efficient, higher accuracy counting models;
\item State-of-the-art counting accuracy is observed on leading benchmarks for several object types including the Shanghaitech \cite{zhang2016single} and Penguins \cite{arteta2016counting} datasets.
\end{itemize}

The remainder of this paper is organised as follows: Section 2 presents the related work found in the literature. Section 3 describes the proposed multi-domain object counting approach while Section 4 details the construction of a challenging new cell counting dataset. Finally, Section 5 presents a comprehensive set of experiments highlighting the development of our technique and the benefits associated with multi-domain object counting.

\section{Related work}
\textbf{Crowd Counting}. Crowd counting has been approached using a wide variety of techniques, from HOG-based head detectors \cite{idrees2013multi} to CNN-based regressors \cite{Zhang2015}. Heatmap-based crowd counting using a fully convolutional neural network was firstly investigated by Zhang \etal \cite{zhang2016single} and subsequently by Marsden \etal \cite{mark_count}, with notable performance gains observed. A novel model switching technique for crowd counting was proposed by Sam \etal \cite{sam2017switching} which firstly classifies the crowd density level of an image region before performing heatmap-based counting using a network which has been optimised for the detected crowd density level. The standard datasets for evaluating crowd counting techniques include the UCF\_CC\_50 \cite{idrees2013multi} and Shanghaitech \cite{zhang2016single} datasets.

\textbf{Cell counting}. Cell nuclei counting techniques have evolved from SIFT-based heatmap estimation \cite{lempitsky2010learning} to fully convolutional neural networks \cite{xie2016microscopy,paul2017count}, with significant improvements in accuracy observed. Significant variation in cell morphology and visual occlusion limit the accuracy of these techniques. The VGG Cell dataset \cite{XieCount} (made up entirely of synthetic images) is the main public benchmark used to compare cell counting techniques. Recently the Modified Bone Marrow (MBM) dataset \cite{paul2017count} was introduced, containing images of real cells observed in histological slides (a 2D cross section of a 3D tissue structure). However, a cell counting dataset in the context of tissue culture and patient diagnosis is still needed.  Cell counting in tissue culture is often done by hand on a daily basis, especially in cases where highly expensive automated systems are not available. A vision-based cell counting system for tissue culture can lead to a more affordable solution and faster diagnoses overall.

\textbf{Vehicle counting}. Vehicle counting has been tackled using a variety of techniques including SIFT-based regression \cite{guerrero2015extremely}, CNN regression \cite{onoro2016towards}, density heatmap generation \cite{kang2017beyond} as well as an LSTM (Long Short-Term Memory) based approach from Zhang \etal \cite{zhang2017fcn}  which adds a temporal dimension to vehicle counting. Vehicle counting techniques are trained and evaluated using the WebCamT \cite{zhang2017fcn} and TRANCOS \cite{guerrero2015extremely} datasets.  

\textbf{Wildlife counting}. Counting of wildlife in ecological studies has not received significant attention from the computer vision community. The core work in this area is that of Arteta \etal \cite{arteta2016counting} who produced a large-scale dataset for counting penguin colonies.

\textbf{Domain adaptation and shared learning models}. While this paper focuses primarily on domain adaptation it also touches on concepts including transfer learning, feature extraction and learning without forgetting (LwF). The ability to train a machine learning model to perform additional  tasks over time while maintaining accuracy in previous tasks has generated significant interest in the computer vision community. This notion is inspired largely by the human visual system, which learns a universal
representation for vision in the early life and uses this representation for a variety of problems \cite{rebuffi2017learning}. Fine-tuning is a common process used to adapt a given neural network to a new task, the main downside being that the original function is often lost during optimisation. Multi-Task learning (MTL) approaches attempt to train a  model to simultaneously perform several tasks, often within a \textit{specific} visual domain. MTL approaches typically involve including sets of task specific layers at end of a given neural network and jointly training to perform all tasks. This type of approach is cumbersome to extend to new tasks and visual domains as all tasks must be retrained.

While intra-domain, multi-objective learning has been utilized heavily to extend the functionality of neural networks, training a model to perform tasks in various \textit{distinct} visual domains (CCTV scenes, medical imaging) has  proven to be more challenging due to the observed variation in statistical distributions between domains.  These issues have been addressed by Bilen \etal \cite{bilen2017universal} who trained a CNN model to learn a universal vision representation which can jointly perform non-related tasks from distinct visual domains by including domain-specific scaling and normalisation layers throughout the network. This work was extended by Rebuffi \etal \cite{rebuffi2017learning} who added domain-specific convolutions and proposed a sequential training procedure for learning new tasks over time without discarding the previously learned functions.

\section{Our approach}

The proposed object counting technique consists of a patch-based CNN regressor with significant inter-domain parameter sharing that can be quickly switched between a learned set of visual domains by interchanging a subset of domain specific parameters.  The  estimated object count for a given image is calculated by summing the object count for all patches. A patch based approach to counting enables a more robust regressor to be learned as objects within a smaller image region are observed to be approximately uniform in size. A lightweight fully convolutional neural network is then used to refine a set of patch estimates for a given image by including the context of adjacent patch estimates to mitigate possible errors.

\subsection{Base object counting regressor}
\begin{figure*}[h!]
\centering

\includegraphics[width=0.9\textwidth]{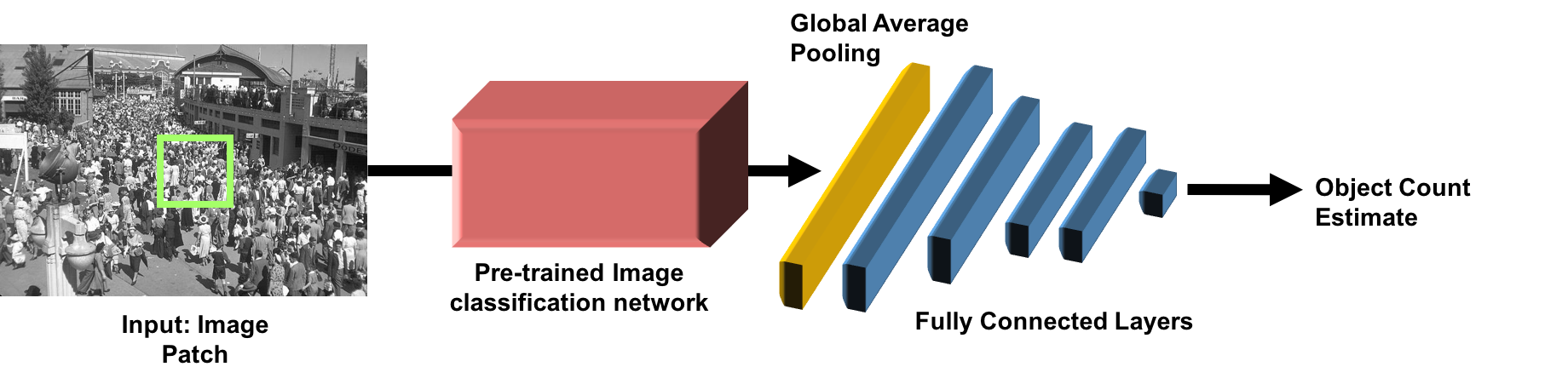}

\caption{Network architecture for the proposed patch-based object counting regressor}
\label{network_viz}
\end{figure*}

The base object counting regressor (shown in Figure \ref{network_viz}) consists of two distinct steps. First a set of high-level features are extracted from each image patch using a pre-trained image classification network, the parameters of which are frozen during model training. The $N$ feature maps generated by the given image classification network's final convolutional layer are average pooled globally to produce an $N$-dimensional feature representation. This $N$-dimensional feature representation is then mapped to an object count value using a fully connected neural network. This fully connected network consists of 5 layers with the following configuration of neurons 256-128-64-64-1. Rectified linear unit (ReLU) activations are applied after each fully connected layer. The trainable portion of the overall network consists of just 330,000 parameters. A variety of pre-trained object classification networks and image patch sizes are investigated in Section 5.

\subsection{Domain-specific layers}

\begin{figure}[h!]
\centering

\includegraphics[width=0.4\textwidth]{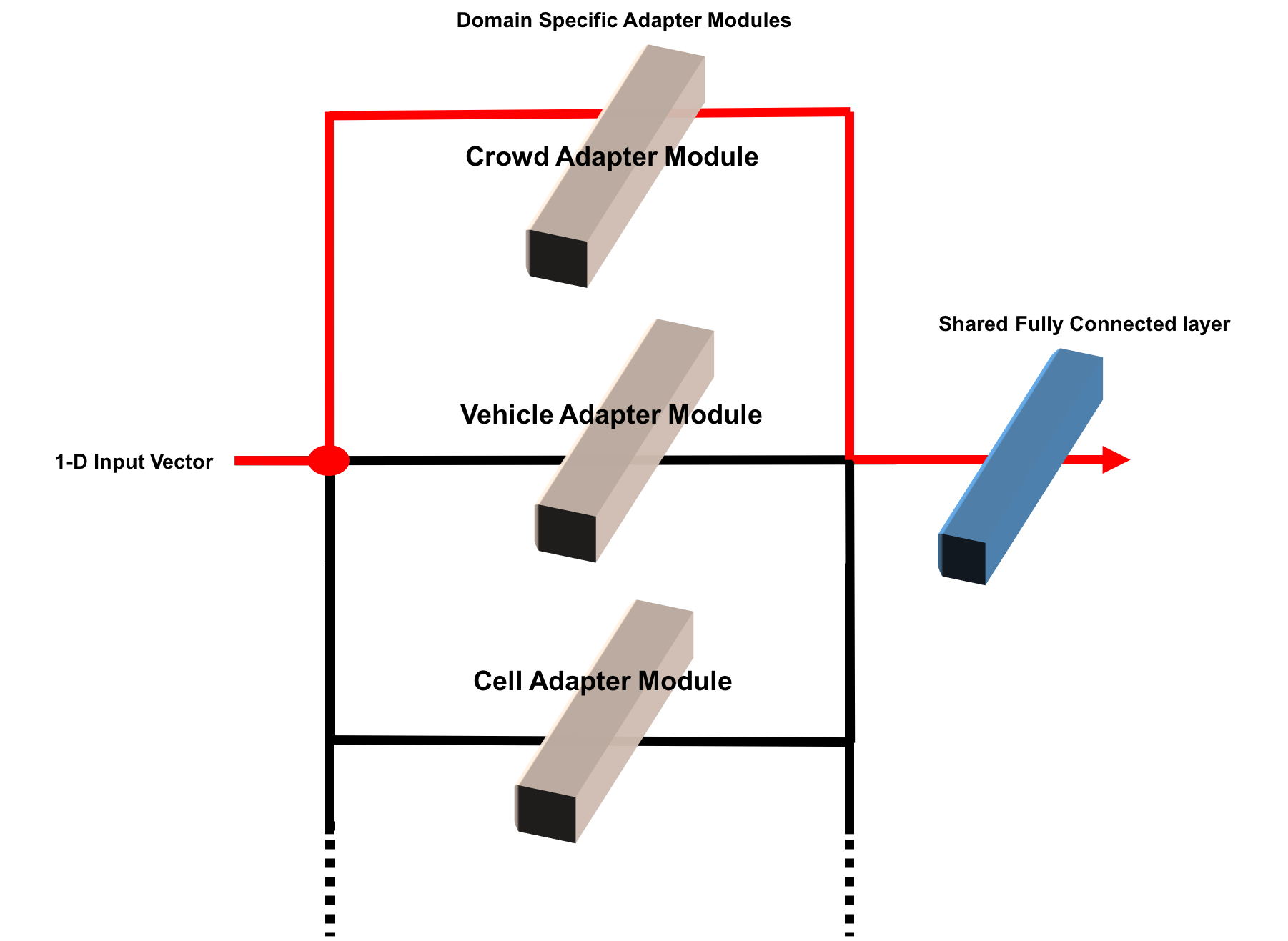}

\caption{Domain specific modules are interchanged during training and inference depending on the chosen counting task (red path).}
\label{switching_concept}
\end{figure}

\begin{figure}[h!]
\centering

\includegraphics[width=0.4\textwidth]{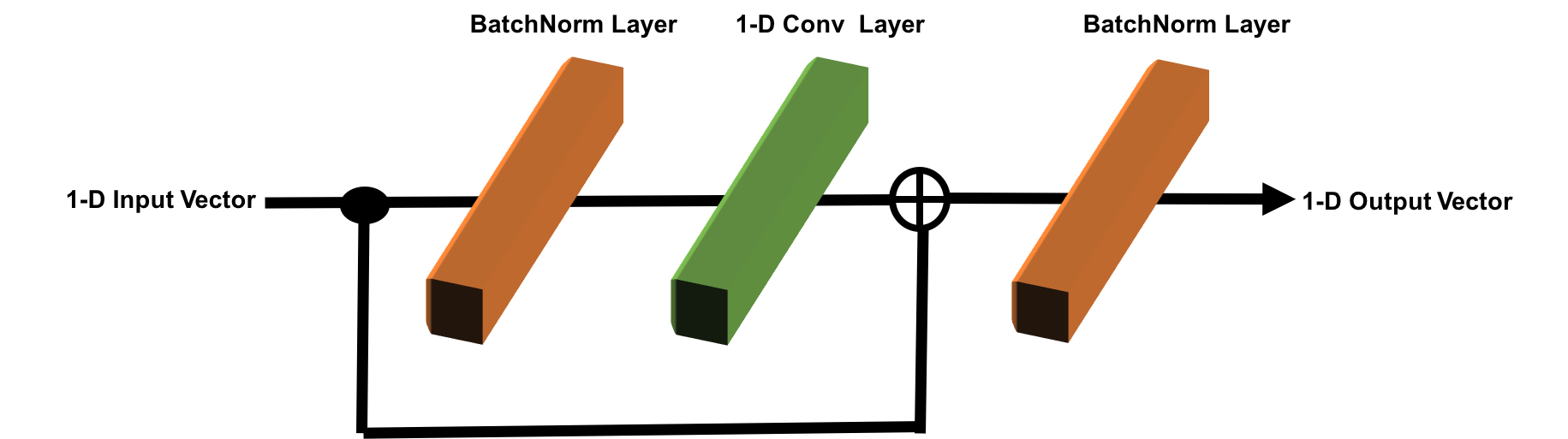}

\caption{The residual adapter module of Rebuffi \etal \cite{rebuffi2017learning}.}
\label{domain_adapter}
\end{figure}

To enable the proposed counting regressor to adapt to various visual domains (e.g. people, vehicles, wildlife, cell nuclei) a set of domain-specific modules are included before each fully connected layer and then after the final fully connected layer. These domain-specific modules are interchanged during training and inference depending on the chosen visual domain (this switching concept is highlighted in Figure \ref{switching_concept}). Each domain-specific module contains the residual adapter of Rebuffi \etal \cite{rebuffi2017learning}  (shown in Figure \ref{domain_adapter}). These residual adapter modules allows for the network to adapt to the distinct statistical distributions of the various visual domains through domain-specific normalisation and scaling \cite{ioffe2015batch}. Including these domain-specific modules throughout the network increases the trainable parameter count by just 5\% for each additional visual domain learned.

\subsection{Sequential training}

A sequential training procedure inspired by the work of Rebuffi \etal \cite{rebuffi2017learning} is used to allow the proposed counting regressor to be extended to new domains over time while still performing the original set of counting tasks. First the network is primed by training to convergence on a given visual domain (e.g. crowd counting), after which the domain-agnostic parameters (i.e. the fully connected layers) are frozen and only the 5\% subset of domain-specific modules are trained to convergence for each of the remaining visual domains, one at a time. Freezing the domain-agnostic parameters enables the network to retain the original counting regression function it has learned. Euclidean distance, given in equation 1 is optimised during training. $\Theta $ corresponds to the set of  network parameters to optimise, \textit{N}  is the batch size, $X_{i}$ is the $i^{th}$  batch image while $F_{i}$ is the corresponding ground truth count value. $F(X_{i};\Theta)$ is the estimated count value for a given batch image $X_{i}$.

\begin{equation}
L_{l2}(\Theta )=\frac{1}{2N}\sum_{i=1}^{N}\left \| F(X_{i};\Theta)-F_{i} \right \|_{2}^{2}.
\end{equation}

The AdaGrad optimiser \cite{duchi2011adaptive} is used to avoid learning rate selection issues with the initial learning rate set to  $1 \times 10^{-1}$. $L_2$ weight regularisation (i.e. weight decay) is also included during model training with $\lambda$ set to $1 \times 10^{-3}$. Model weights are initalised using the uniform initaliser of Glorot and Bengio \cite{glorot2010understanding} while the bias terms are initalised to zero. Training is carried out for 10,000 iterations for each domain. The choice of visual domain used to prime the network is investigated in section 5.

\subsection{Fully convolutional refinement network}
\begin{figure}[h!]
\centering

\includegraphics[width=0.4\textwidth]{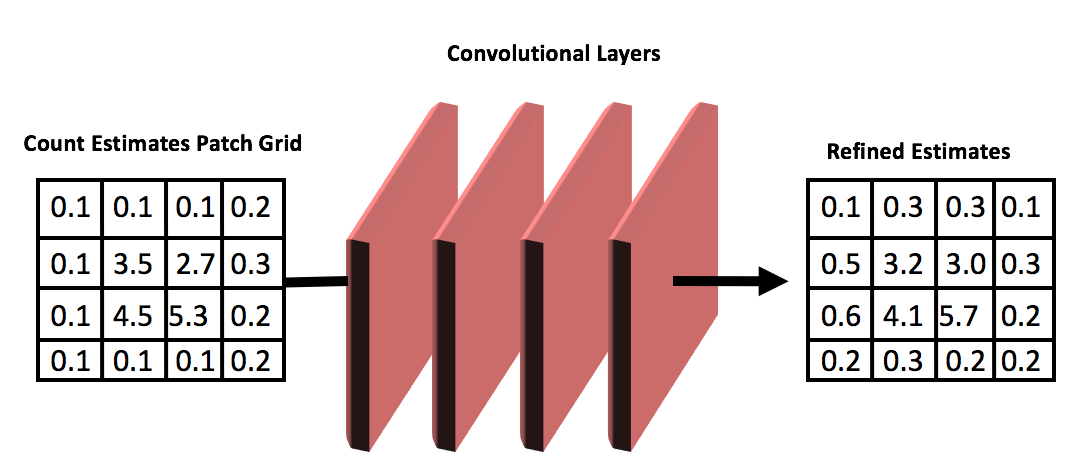}

\caption{Fully convolutional network used for count estimation refinement.}
\label{refinement_net}
\end{figure}

A patch-based counting regressor does not include the wider scene context when processing a given image as patches are analysed on an individual basis. To address this, a fully convolutional neural network is trained to refine a grid of patch estimates for a given image. The proposed fully convolutional network, shown in Figure \ref{refinement_net}, consists of 4 convolutional layers with the following number of kernels per layer: 16,16,16,1. Rectified linear unit (ReLU) activations are applied after each layer. All convolutional kernels are  $3 \times 3$ resulting in a total parameter count of 4950. 

A refinement model can be trained for the count regressor of a given domain by firstly producing a grid of patch estimates for each training image used as well as a corresponding ground truth grid. Training is then carried out using these grid pairs for 10,000 iterations with Euclidean distance again minimised.  Once trained, a refinement model can be applied to a set of count estimates of any width and height due to the fully convolutional nature of the model. During inference this step results in a near negligible increase in processing time.

\subsection{Domain classification}

If the visual domain observed during inference is unknown this can be predicted by extending the core network to also perform  domain classification. To accomplish this the final fully connected layer is interchanged with a $K$-neuron fully connected layer, where $K$ is the number of visual domains to distinguish between. Following this final layer a softmax activation is applied and the subsequent domain adapter module is not included. Training can then be performed by freezing the common model parameters (the initial 4 fully connected layers) and creating a fresh set of adapter modules. Categorical cross-entropy loss, defined as:

\begin{equation}
L_{\text{CCE}}(\Theta )=-\frac{1}{N}\sum_{i=1}^{N} \sum_{j=1}^{K} S_{ij}\log(\hat{S}_{ij}),
\label{den_CCE}
\end{equation}

\noindent is then minimized. $\Theta $ corresponds to the set of trainable network parameters, $N$ and $K$  are the batch size and number of visual domains to be distinguished between while $\hat{S}_{ij}$ is the predicted probability score for concept $j$ in the $i^{th}$  batch image and ${S}_{ij}$ is the corresponding ground truth value.

\section{Dublin cell counting dataset}

When considering the application of computer vision to tissue culture and patient diagnosis there is a  clear lack of publicly available and fully annotated cell microscopy datasets. The main dataset used to evaluate techniques for this task consists entirely of synthetic images \cite{xie2016microscopy}. To address this,  the Dublin Cell Counting (DCC) dataset was constructed.

This dataset consists of 177 images containing a wide array of tissues and species. Amongst these are
examples of stem cells derived from embryonic mice, isolated human lung
adenocarcinoma and examples of primary human monocytes isolated from a healthy human volunteer. Several factors were varied during image capture to provide a more representative set of images. First, the density of cells loaded onto the slide naturally varies as cell lines proliferate at different rates. Second, the morphology and size of
the cells for each cell line can vary significantly. Furthermore,  the objective lens used during imaging was varied as was the diameter of the diaphragm which controls the
amount of light hitting the sample. Finally,
the haemocytometer grid size was varied to produce a representative set of non-cellular image artifacts. Cell images were obtained via a camera mounted on an Olympus CKX41
microscope using both $4\times$ and $10\times$ objectives. The high levels of variation in this collection allow for a more robust cell counting function to be learned. After the full set of image were acquired, a dot annotation process was performed by a domain expert with a background in molecular biology.  

\begin{figure}

\begin{minipage}{\textwidth}

\begin{subfigure}{\textwidth}
\includegraphics[width=0.23\textwidth]{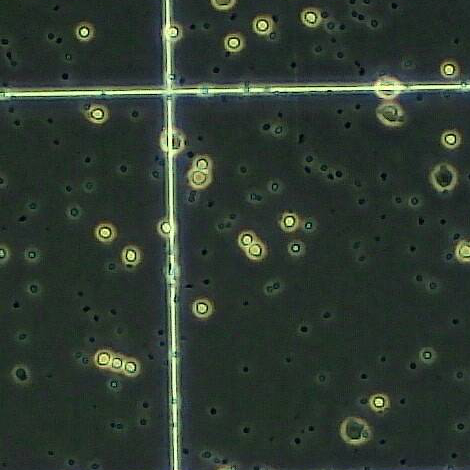}
\space
\includegraphics[width=0.23\textwidth]{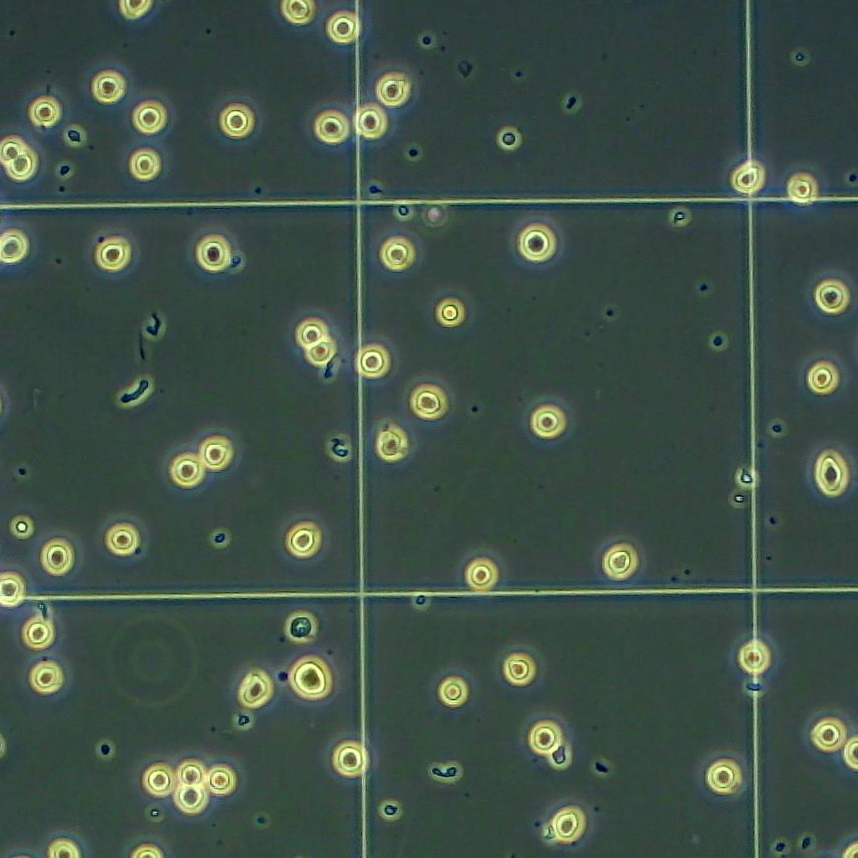}
 
\end{subfigure}
\hfill
\begin{subfigure}{\textwidth}

\vspace{0.1cm}

\includegraphics[width=0.23\textwidth]{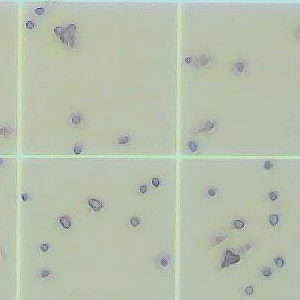}
\space
\includegraphics[width=0.23\textwidth]{cell_examples/DC_3}
\end{subfigure}
\end{minipage}
\hfill
\caption{DCC dataset examples showing the significant variation within this collection.}
\label{cell_examples}
\end{figure}

The mean cell count across these images is 34.1 with a standard deviation of 21.8, showing the significant variation in cell density. 100 images are used for training and validation while the remaining 77 form an unseen test set. Sample images from this collection are shown in Figure \ref{cell_examples}.

\section{Experiments}

\begin{table*}[!htbp]
\centering
\begin{tabular}{ m{2.5cm} m{4.0cm} m{2.0cm}  m{1.5cm} m{1.5cm} m{1.5cm} m{1.5cm}}
\hline
Visual Domain & Dataset & No. of Images  & Count Mean & Count STD & Count Range \\ \hline
Crowd & Shanghaitech (Part A) \cite{zhang2016single} & 482  & 501.4 & 456.4 &  33-3139  \\ 
Vehicles & TRANCOS \cite{guerrero2015extremely} & 1641  &   36.54 &  14.9 & 9-95 \\ 
Wildlife & Penguins \cite{arteta2016counting} & 80095 &  7.18   & 5.71 & 0-67 \\ 
Cells & Dublin Cell Counting  & 177 &  34.1 & 21.8 &0-101  \\ \hline
\end{tabular}
\caption{Details of the datasets used to evaluate the proposed multi-domain object counting technique.}
\label{dataset_contents}
\end{table*}

The proposed multi-domain object counting technique is evaluated using a challenging and representative dataset for each  visual domain including the proposed DCC dataset. These collections are detailed in full in Table \ref{dataset_contents}.  Horizontal flips are used for  training set augmentation in all cases. A 30\% subset of the provided training data is set aside as a validation set for all model selection experiments, apart from where an explicit validation set is provided. No count estimate refinement is applied until subsection 5.5.  Counting performance is evaluated for all visual domains using Mean Absolute Error (MAE) and Root Mean Squared Error (MSE), which are defined as follows:

\begin{equation}
\text{MAE}=\frac{1}{N}\sum_{i=1}^{N}\left |z_i-\check{z}_i  \right |, 
\end{equation}
\begin{equation}
\text{MSE}=\sqrt{\frac{1}{N}\sum_{i=1}^{N}( z_i-\check{z}_i  )^{2} },
\end{equation}

\noindent All network optimisation and testing is performed using an NVIDIA GeForce GTX 970 GPU with a batch size of 256 and implemented using the Keras API \cite{chollet2015keras} with a Tensorflow backend \cite{abadi2015tensorflow}.

\subsection{Patch size selection}

Several image patch sizes (50 $\times$ 50,  100 $\times$ 100, 200 $\times$ 200) are compared by evaluating crowd counting validation performance on the Shanghaitech dataset (part A). Feature extraction for the base counting regressor is performed using the VGG16 network of Simonyan and Zisserman \cite{simonyan2014very} for all runs. A given patch size is used during both dataset construction and inference.  Both the domain-agnostic and domain-specific network parameters are optimised for each run in this experiment as the network has not been primed on any other dataset. Table \ref{patch_performance} highlights the performance of the various patch sizes. 100 $\times$ 100 patches result in the best overall performance and will be used in all subsequent experiments. This size likely strikes the optimal balance between uniform object size and the inclusion of wider scene context needed to learn a robust function.  On the other hand, the use of 200 $\times$ 200 patches results in significantly inferior performance due to the non-uniform object sizes observed in larger image regions.

\begin{table}[!htbp]
\centering
\begin{tabular}{ l l l}
\hline
Patch Size & MAE & MSE \\ \hline 
50 $\times$ 50 & 100.8 & 152.5 \\ 
100 $\times$ 100 &  \textbf{97.5} & \textbf{145.4} \\ 
200 $\times$ 200 & 158.5 & 247.3 \\ \hline
\end{tabular}
\caption{Crowd counting validation performance of various image patch sizes on a validation set taken from the Shanghaitech (Part A) dataset.}
\label{patch_performance}
\end{table}

\subsection{Feature extractor selection}
Several pre-trained image classification networks are evaluated as feature extractors for object counting. These networks include the ResNet50 network of He \etal \cite{he2016deep} the VGG16 network of Simonyan and Zisserman \cite{simonyan2014very} as well as the  MobileNet architecture of Howard \etal \cite{howard2017mobilenets}. Crowd counting validation performance is again evaluated on the Shanghaitech (Part A) dataset with a patch size of 100 $\times$ 100 used. Table \ref{FE_performance} details the various pre-trained networks and presents the performance achieved by each. The number of  parameters associated with each network does not include the original set of fully connected layers (which have been replaced entirely). The MobileNet feature extractor achieves the best overall performance despite having significantly fewer parameters and is used for all subsequent experiments. The use of this  lightweight architecture also potentially makes the developed method more suitable for edge processing on low power devices.

\begin{table}[!htbp]
\centering
\begin{tabular}{l m{1.5cm} l m{0.75cm} m{0.75cm}}
\hline
Net & Parameters  & MAE & MSE \\ \hline 
VGG16 \cite{simonyan2014very} & 14M  & 101.2 & 151.4 \\ 
Resnet50 \cite{he2016deep} & 24M  & 97.5 & 145.4 \\ 
MobileNet ($\alpha=1.0$) \cite{howard2017mobilenets} & 3.5M  & \textbf{94.1} & \textbf{138.9} \\ \hline
\end{tabular}
\caption{Validation performance on the Shanghaitech dataset (part A) for various image classification networks being used for feature extraction.}

\label{FE_performance}
\end{table}

\subsection{Priming the network}
In the work of Rebuffi \etal \cite{rebuffi2017learning} a multi-domain  image classification network is primed on the ImageNet dataset \cite{imagenet_cvpr09}  due to it's size and variety before being adapted to other domains. However, for the proposed multi-domain object counting model the choice of which domain with which to prime the network is not clear. Therefore all 4 domains (crowds, vehicles, wildlife, cells) are evaluated for this purpose by priming the network from scratch and then adapting to the remaining 3 domains one by one using the sequential training process discussed previously.  Table \ref{sequential_training} presents the MAE score observed for each pairing. Each row in this table corresponds to the domain used to prime the network while each column corresponds to the MAE performance achieved when adapting the model. The diagonal entries correspond to the performance achieved when training the network from scratch for each domain. 

It can be seen that priming the network on the cell domain using the DCC dataset results in the best overall performance, achieving optimal MAE on 3 of the 4 domains. Adapting from the cell domain achieves performance superior to training from scratch on both the crowd and wildlife domains, which is noteworthy given the small number of domain-specific parameters
 trained (5\%). The high performance observed across domains  when adapting from a cell counting model is likely due to the significant morphological variation observed between cell objects in the DCC dataset, resulting in a broader set of learned features. In all subsequent experiments the counting network is primed on the cell counting task.

\begin{table*}[!htbp]
\centering
\begin{tabular}{l l l l l l l l}
\hline
Visual Domain & Crowd (Tested) & Vehicles (Tested) & Wildlife (Tested) & Cells (Tested) & Best Overall  \\ \hline 
Crowd (Primed)         & 94.1     & 10.3        &  6.8     & 12.9     &  0/4   \\ 
Vehicles (Primed)      &  96.2    &  9.9      &  6.1    &   11.5    &  1/4  \\ 
Wildlife (Primed)      & 95.3     &   10.3     & 6.05  &  10.2   &    0/4  \\ 
Cells  (Primed)       & \textbf{90.2}    &   10.2      &  \textbf{5.7}    &  \textbf{9.5}    &    \textbf{3/4} \\ \hline
\end{tabular}
\caption{The MAE validation performance achieved when priming the network using each of the 4 visual domains.}
\label{sequential_training}
\end{table*}

\subsection{Comparison with feature extraction based network adaptation}
Feature extraction based network adaptation involves freezing the majority of a given neural network and retraining the final few layers for a new task (with the task-specific parameters then interchanged as required). This approach retains the original function but often requires a large number of new parameters to be trained. In this experiment we compare the Rebuffi domain adapter \cite{rebuffi2017learning} with feature extraction as a domain adaptation strategy. The approaches are evaluated in terms of MAE performance in the target domain and the quantity of new model parameters introduced. The number of retrained fully connected layers is varied to investigate the difference in performance. All adapter modules placed before trainable layers are re-trained while those before frozen layers are also frozen. A given cell counting model is adapted to perform crowd counting on the Shanghaitech dataset (part A) in each case.  Figure \ref{perf_domain_adapt} highlights the superior performance of the Rebuffi domain adapter when adapting to crowd counting, despite using significantly fewer parameters.

\begin{table}[h!]
\centering
\begin{tabular}{lll}
\hline
Approach  & Extra Parameters & MAE \\ \hline
From Scratch  & 330K & 94.1 \\
Final 4 layers re-trained  & 50K & 128.3 \\
Final 3 layers re-trained  & 15K & 232.2 \\
Final 2 layers re-trained & 5K & 245.5 \\
Rebuffi \etal \cite{rebuffi2017learning} & 16K & \textbf{90.2} \\ \hline
\end{tabular}
\caption{MAE validation performance on the Shanghaitech dataset (part A) for various domain adaptation strategies.}
\label{perf_domain_adapt}
\end{table}

\subsection{Prediction refinement}
\begin{figure*}
\captionsetup[subfigure]{justification=centering}
    \centering
      \begin{subfigure}{0.233\textwidth}
        \includegraphics[width=\textwidth]{test/IMG_5}
          \caption{Domain: Crowd \\ Predicted: 801.3 GT: 819}
          \label{fig:NiceImage1}
      \end{subfigure}
      \space
      \begin{subfigure}{0.23\textwidth}
        \includegraphics[width=\textwidth]{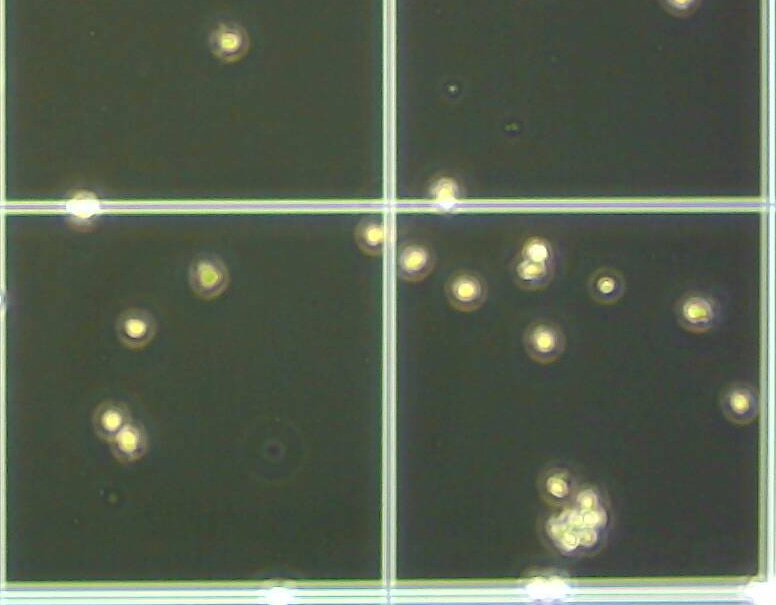}
          \caption{Domain: Cells \\ Predicted: 25.6 GT: 25}
          \label{fig:NiceImage2}
      \end{subfigure}
      \space
      \begin{subfigure}{0.23\textwidth}
        \includegraphics[width=\textwidth]{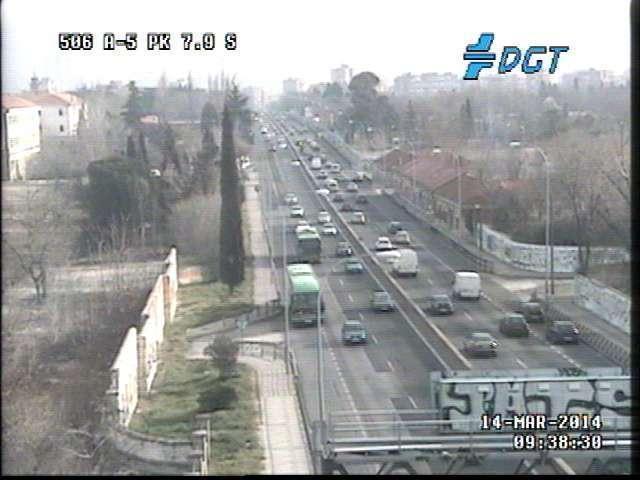}
          \caption{Domain: Vehicles \\ Predicted: 29.2 GT: 29}
          \label{fig:NiceImage3}
      \end{subfigure}
      \space
       \begin{subfigure}{0.233\textwidth}
        \includegraphics[width=\textwidth]{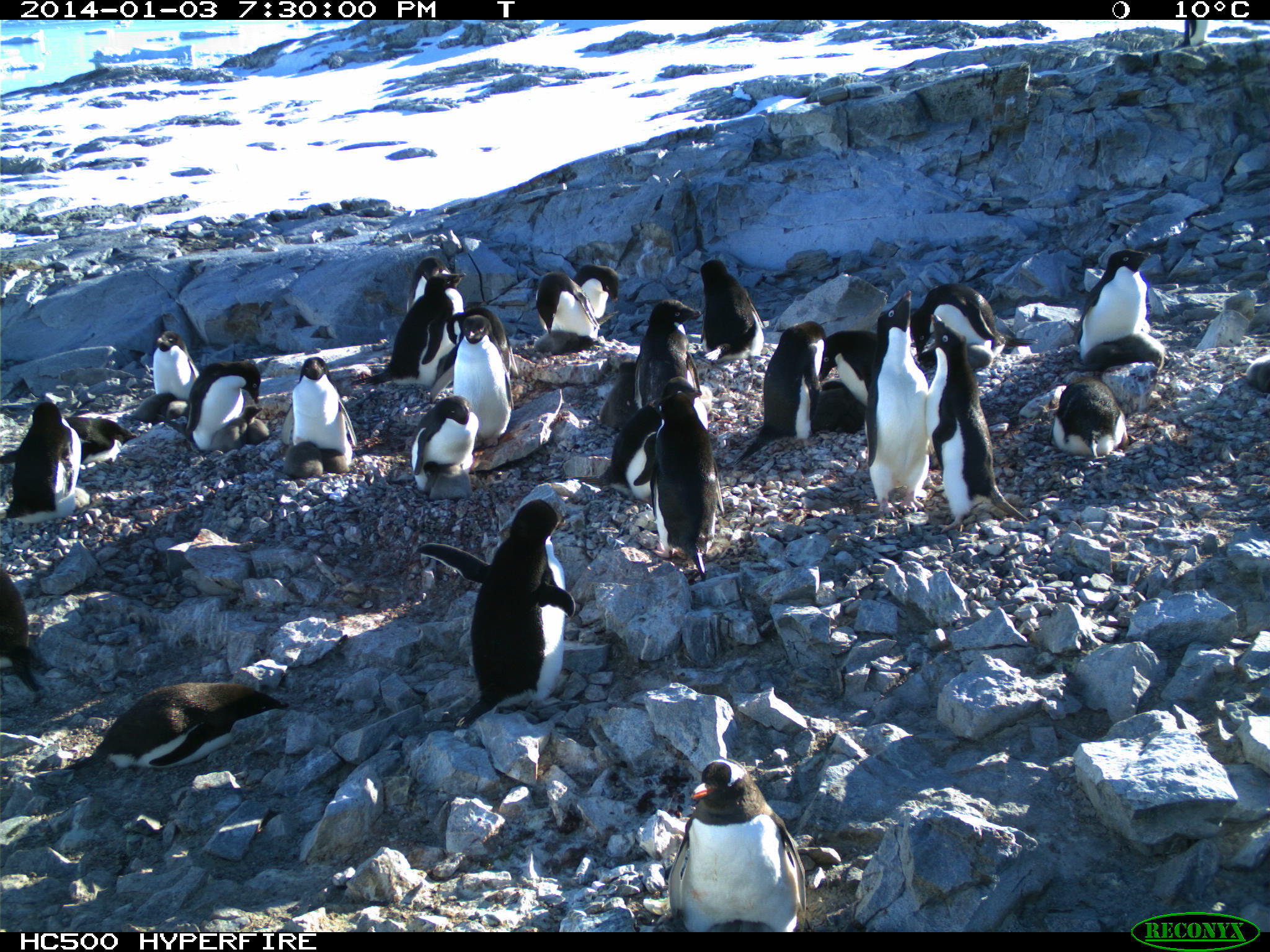}
          \caption{Domain: Wildlife \\ Predicted: 25.2 GT: 26}
          \label{fig:NiceImage3}
      \end{subfigure}
      \caption{The robust performance  of the proposed counting model is highlighted on images taken from each of the 4 visual domains used in this study.}
     \label{test_examples}
\end{figure*}

A  fully convolutional refinement model is trained for each of the 4 visual domains and compared to the validation performance of the base regressor model in each case. Results of this experiment are shown in table \ref{refinement_results}. This efficient post-processing step has a near negligible impact on inference time but results in significant performance boosts in the crowd and cell counting benchmarks. Examples of the final count regressor in action across all 4 domains are shown in figure \ref{test_examples}.

\begin{table}[!htbp]
\centering
\begin{tabular}{l l l}
\hline
Visual Domain &  Base MAE & Refined-MAE\\ \hline
Crowds  &90.2 &86.5  \\
Vehicles & 10.2  & 10.1 \\ 
Wildlife & 5.7  & 5.6  \\ 
Cells & 9.5 & 8.4  \\ \hline
\end{tabular}
\caption{The MAE observed on the validation sets of all 4 visual domains for the base regressor and after the proposed refinement step has been applied.}
\label{refinement_results}
\end{table}

\subsection{Domain classification performance}
A domain classifier is trained by firstly producing a representative dataset taken from all 4 visual domains used in this study.  300 full scene images are taken from each collection and a 6:4 training/test split is then applied, ensuring an even number of images are taken from each domain. Horizontal image flips are applied to produce a 300 image set from the smaller DCC dataset. Table \ref{domain_classification} compares classifier accuracy when training from scratch, when adapting from a cell counting network using a fresh set of adapter modules and when adapting from a cell counting network and just training the final fully connected layer.

Near perfect domain classification accuracy is observed when training the network from scratch and also when adapting from the a cell counting network using  a new set of adapter modules. However, when we only train the final fully connected layer to perform classification we observe significant performance degradation,  showing the importance of the included domain adapter modules despite their low parameter count.

 \begin{table}[!htbp]
\centering
\begin{tabular}{ l l l l}
\hline
Primer &Training Approach & Accuracy  \\ \hline 
None & Entire Network Trained & \textbf{99.7\%}\\ 
Cells &Adapter Modules + Final Layer & 98.2\% \\ 
Cells &Final Layer Only  & 56.3\% \\  \hline
\end{tabular}
\caption{Domain classification accuracy as the training approach is varied and domain adapter modules are included.}
\label{domain_classification}
\end{table}

\subsection{Comparison with the state-of-the-art}
The developed object counting technique is compared to the leading techniques across the 4 visual domains used in this study. Evaluation is performed in all cases on a previously unseen test set. The network is primed on cell counting using the DCC dataset. The proposed refinement step is applied for each domain. Table \ref{perf_SHT} compare crowd counting performance on the Shanghaitech dataset (parts A and B), with state-of-the-art performance achieved on both. The commonly used UCF\_CC\_50 dataset \cite{idrees2013multi} has been deemed inappropriate for benchmarking as it contains crowds too dense for the human eye to count \cite{Hu2016} and is therefore not used to evaluate the proposed technique. Table \ref{perf_TRANCOS} compares the proposed technique with the leading vehicle counting techniques on the TRANCOS dataset while table \ref{perf_MBM} highlights cell counting performance on the Modified Bone Marrow dataset \cite{cohen2017count}. Finally, Table \ref{perf_PENG} shows the  superior performance of our technique on the Penguins dataset \cite{arteta2016counting}.

\begin{table}[t]

\centering
\begin{tabular}{l l l l l }
\hline
                                   & \multicolumn{2}{c}{Part A}         &  \multicolumn{2}{c}{Part B}     \\ \hline  
Method                             & MAE            & MSE           & MAE           & MSE           \\ \hline 
Zhang \etal \cite{zhang2016single}                 & 110.2          & 173.2          & 26.4          & 41.3          \\ 
Marsden \etal \cite{mark_count}                       & 126.5 & 173.5 & 23.76 & 33.12 \\ 
Switch-CNN  \cite{sam2017switching}                       & 90.4 & 135.0 & 21.6 & 33.12 \\ 
Our-Approach                        &\textbf{85.7 }& \textbf{131.1} &\textbf{17.7} & \textbf{28.6} \\ \hline
\end{tabular}
\caption{Comparing the performance of various crowd counting approaches on the Shanghaitech dataset.}
\label{perf_SHT}
\end{table}

\begin{table}[t]
\centering
\begin{tabular}{l l l}
\hline
Method                             & MAE                         \\ \hline 
Hydra-CNN \cite{onoro2016towards}                      & 10.99 \\ 
Our Approach                       & 9.7 \\ 
Zhang \etal \cite{zhang2017fcn}                      & \textbf{4.2} \\ \hline
\end{tabular}

\caption{Comparing performance of various vehicle counting approaches on the TRANCOS dataset. }
\label{perf_TRANCOS}
\end{table}

\begin{table}[t]
\centering
\begin{tabular}{ l l l l}
\hline
Method & N=5 & N=10 & N=15 \\ \hline
\cite{xie2016microscopy} & 28.9 $\pm$ 22.6 & 22.2$\pm$ 11.6 & 21.3$\pm$ 9.4 \\ 
Ours & 23.6$\pm$ 4.6 & 21.5$\pm$ 4.2 & 20.5$\pm$ 3.5 \\ 
\cite{cohen2017count}  & \textbf{12.6$\pm$ 3.0} & \textbf{10.7$\pm$ 2.5} & \textbf{8.8$\pm$2.3} \\ \hline
\end{tabular}
\caption{Cell counting MAE performance on the MBM dataset. Out of the 44 images in this collection, N are used for training, N for validation and  an unseen 14 images for testing. At least 10 runs using random dataset splits are performed for the each N value.}
\label{perf_MBM}
\end{table}

\begin{table}[t]
\centering
\begin{tabular}{l l}
\hline
Method                             & MAE                           \\ \hline 
Arteta \etal \cite{arteta2016counting}                     & 8.11  \\ 
Our Approach                       & \textbf{5.8}  \\ \hline

\end{tabular}

\caption{Comparing performance of various counting techniques on the Penguins dataset test set. MAE is computed w.r.t the max count on each image (as there are multiple annotators). The separate site dataset split is used and no depth information is utilised.}
\label{perf_PENG}
\end{table}

\section{Conclusion}
In this paper we propose a new multi-domain object counting technique that employs significant parameter sharing and achieves state-of-the-art benchmarking performance for several visual domains. This model can be extended to new counting tasks over time while still maintaining identical performance in all prior tasks. The benefits of this singular approach to object counting include the removal of redundant model parameters as well as noticeable increases in counting accuracy over single-domain baseline runs. Each new counting task requires just 20,000 additional model parameters (including the proposed estimate refinement step). The Dublin Cell Counting (DCC) dataset was introduced as part of this study. This new collection is the first of it's kind and contains a challenging and highly-varied set of cellular images in the context of tissue culture and patient diagnosis. Using the developed single-model approach state-of-the-art object counting performance was observed in the Shanghaitech dataset (parts A and B) as well as the Penguins dataset. Future work in this area will look to extend the proposed counting model to perform additional regression and classification tasks in fields including medical imaging, crowd behaviour analysis and scene understanding.

\section*{Acknowledgements}
This publication has emanated from research conducted with the financial support of the Irish Research Council and Science Foundation Ireland (SFI) under grant numbers SFI/12/RC/2289 and 15/SIRG/3283. A special thanks is also given to the Cummin's group in the School of Medicine, University College Dublin for making the DCC dataset images available to the wider research community.

{\small
\bibliographystyle{unsrt}
\bibliography{egbib}
}

\end{document}